\UseRawInputEncoding
\documentclass[letterpaper, 10 pt, conference]{ieeeconf}  

\IEEEoverridecommandlockouts                              

\usepackage{longtable}
\usepackage{booktabs}
\usepackage{algorithm}
\usepackage{bbm}
\usepackage{xcolor}
\usepackage{multirow}
\usepackage{amsfonts}
\usepackage{amsmath,amssymb,amsfonts}
\usepackage{algorithmic}
\usepackage{graphicx}
\usepackage{textcomp}
\usepackage{makecell}
\usepackage{cite}
\usepackage{amsmath}
\usepackage{amssymb}
\usepackage{multicol}
\usepackage{bm}
\usepackage{array}
\usepackage[hidelinks]{hyperref}

\usepackage{siunitx} 
\begin{document}
\title{\LARGE \bf
Dynamic SpectraFormer for Ultra-High-Definition Underwater Image Enhancement{  
}}

\author{Zhiqiang HU$^{1}$, Tao YU$^{2}$, Shouren HUANG$^{1}$ and Masatoshi ISHIKAWA$^{1}$
\thanks{}
\thanks{$^{1}$Zhiqiang HU, Shouren HUANG and Masatoshi ISHIKAWA are with Research Institute for Science \& Technology, Tokyo University of Science
        {\tt\small \{zhiqiang.hu, huang, ishikawa\}@ishikawa-vision.org}}%
\thanks{$^{2}$Tao YU is with Tokyo Institute of Technology
        {\tt\small yutao@mobile.ee.titech.ac.jp
}}%
}

\maketitle
\thispagestyle{empty}
\pagestyle{empty}

\begin{abstract}
Underwater images suffer from color distortion, haze, and poor visibility due to light refraction and absorption in water. These challenges significantly impact the utilization of Autonomous Underwater Vehicles (AUVs) or marine robots. Typically, color and brightness distortions manifest at lower frequencies, while edge and texture distortions are prevalent at higher frequencies. Traditional methods struggle to concurrently rectify these mixed distortions as they primarily concentrate on the spatial domain. To address these issues, we introduce the Dynamic SpectraFormer, which enhances underwater images through a frequency domain transformer. The Dynamic SpectraFormer introduces an ultra-high-resolution sparse spectrum attention module, which could capture the long-term dependency without losing the universal approximating power. Additionally, we have developed a dynamic spectrum weight generation layer that serves as an adaptive spectrum band selector, accentuating critical frequency bands and suppressing less relevant ones. Consequently, this method significantly improves underwater image quality by addressing both high- and low-frequency distortions. Our extensive ablation studies and comparative evaluations consolidate the Dynamic SpectraFormer's efficacy across multiple underwater image enhancement benchmarks. The source code
is available at \href{https://github.com/arifence2024/DynamicSpectraFormer.git}{https://github.com/arifence2024/DynamicSpectraFormer.git}.	
\end{abstract}
\section{Introduction}
\label{sec:intro}
The degradation in underwater image quality hampers the visual sensing capabilities of marine robots, despite that they are equipped with high-end cameras. Thus, algorithms for Underwater Image Enhancement (UIE) play a critical role in advancing aquatic exploration, with widespread applications in domains like Autonomous Underwater Vehicles (AUVs) and Remotely Operated Vehicles (ROVs). This degradation in image quality is primarily due to the wavelength-dependent scattering and attenuation of light as it travels through water. For instance, red light, having the longest wavelength, is absorbed first, followed by green and then blue light. 
\begin{figure}[tbp]
	\centering
	\includegraphics[width=1\linewidth]{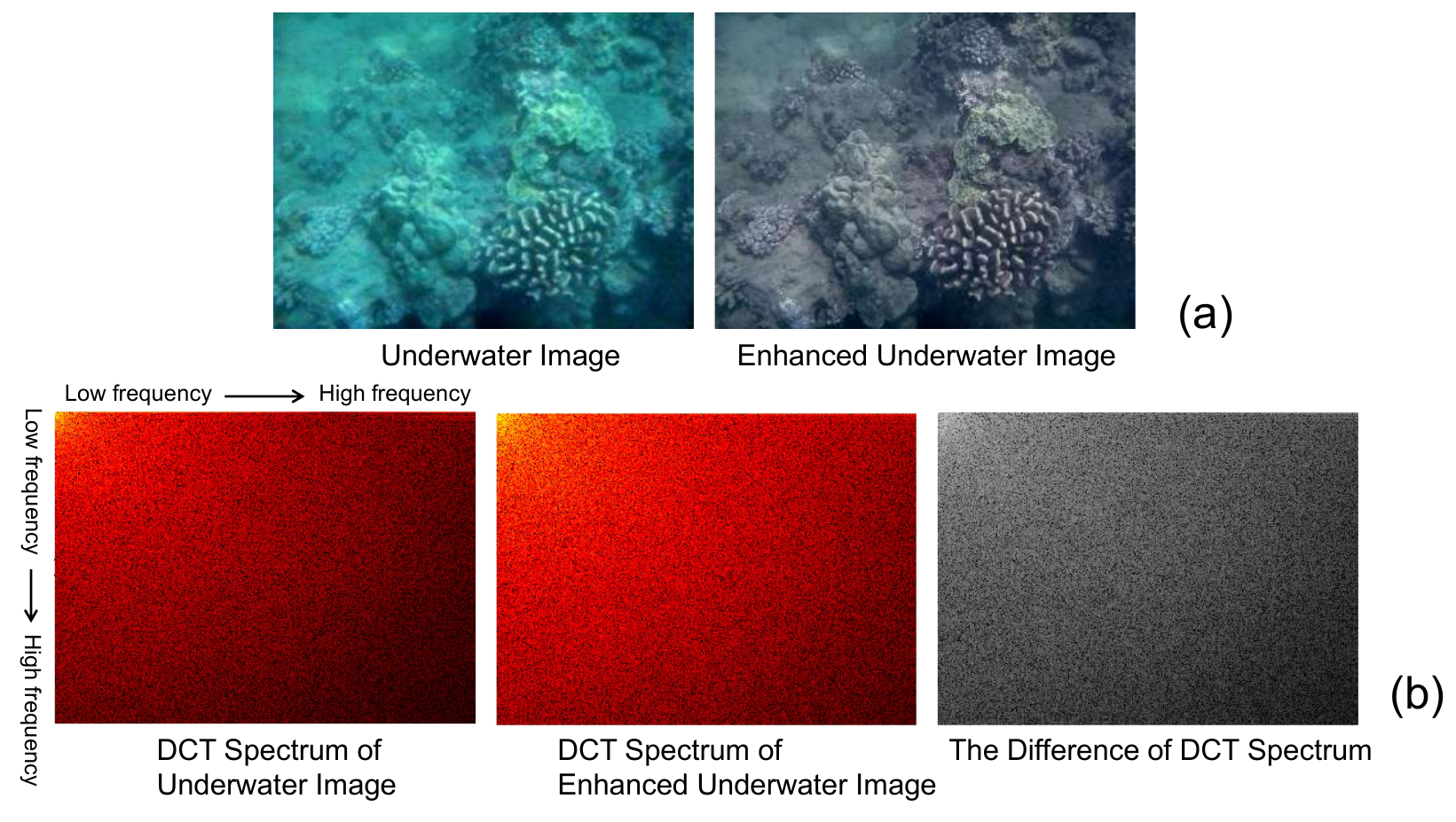}
	\caption{The difference of the DCT spectrum between a pair of underwater image and its enhanced image in LSUI  \cite{peng2023u} dataset. Where (a) are the image pairs, (b) the corresponding DCT spectrum and the difference. The enhanced image contains more high frequencies compared with the original image.}
 \vspace{-2mm}
\end{figure}
The enhancement of Ultra High Definition (UHD) underwater imagery faces challenges from two primary aspects: Initially, underwater images often contain spatially variable hybrid degradations, predominantly at high frequencies, such as blurred textures. Additionally, varied water types manifest unique distortion characteristics primarily at low frequencies, such as color distortion, haze, etc. Moreover, enhancing UHD images is also a computationally intensive task, which hinders its application on devices with limited resources.
\begin{figure*}[htbp]
	\centering
	\includegraphics[width=0.9\linewidth]{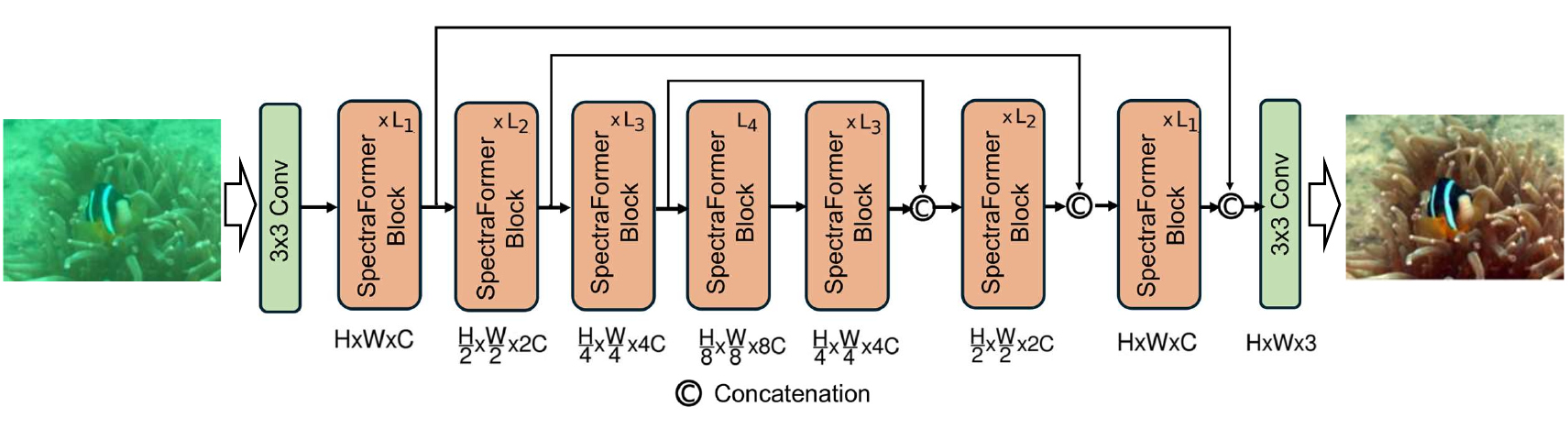}
	\caption{Dynamic SpectraFormer is configured with a multi-scale U-shape architecture for restoring high-definition underwater images.}
\end{figure*}
To tackle these challenges, studies \cite{Khan_2024_WACV, sharma2023wavelength, wei2022uhd} focus on enhancing underwater image quality through the frequency domain. Typically, these approaches leverage Fourier or wavelet transforms to extract frequency domain coefficients from images that have deteriorated. Following this, various techniques like thresholding, filtering, or deep learning are employed to minimize the differences between the degraded and clear images. However, these approaches have an inherent limitation: they treat the spectrum uniformly across all frequency bands without considering the unique characteristics of the input. We argue that the image-agnostic global filter is not the optimal choice. This disadvantage limits their ability to adjust to diverse content types and often results in inadequate performance, especially in handling complex image distortions in underwater circumstances. 

On the other hand, deep learning approaches \cite{Yang2020cGAN, hu2022data, Li2027WaterGAN, Jiang2020Perceptual, Li2026WaterNet, Hu_2022_ACCV} have shown outstanding effectiveness in the domain of visual perception and enhancement. Recent studies, including those by \cite{peng2023u}, \cite{ren2022reinforced}, and \cite{Khan_2024_WACV}, have shifted focus to employing Vision Transformers (ViTs) \cite{dosovitskiy2020image} for enhancing underwater images. ViTs are known for their effective handling of complex patterns over large image areas. However, for the image of \( W\times H \) pixels, the time and memory complexity of the key-query dot-product interaction increase quadratically with the spatial resolution of input, that is, \( \mathcal{O}(W^2H^2) \). Consequently, applying pixel-level self-attention to restore the high-resolution images is not possible. To mitigate this, the self-attention with local-window techniques \cite{peng2023u, ren2022reinforced} is proposed to enhance the image quality, however, such methods can hardly capture extensive spatial information of the whole image and introduce noticeable defects in the restored images. This limitation inherently restricts its application in UHD UIE tasks.
\begin{figure*}[tbp]
	\centering
	\includegraphics[width=0.88\linewidth]{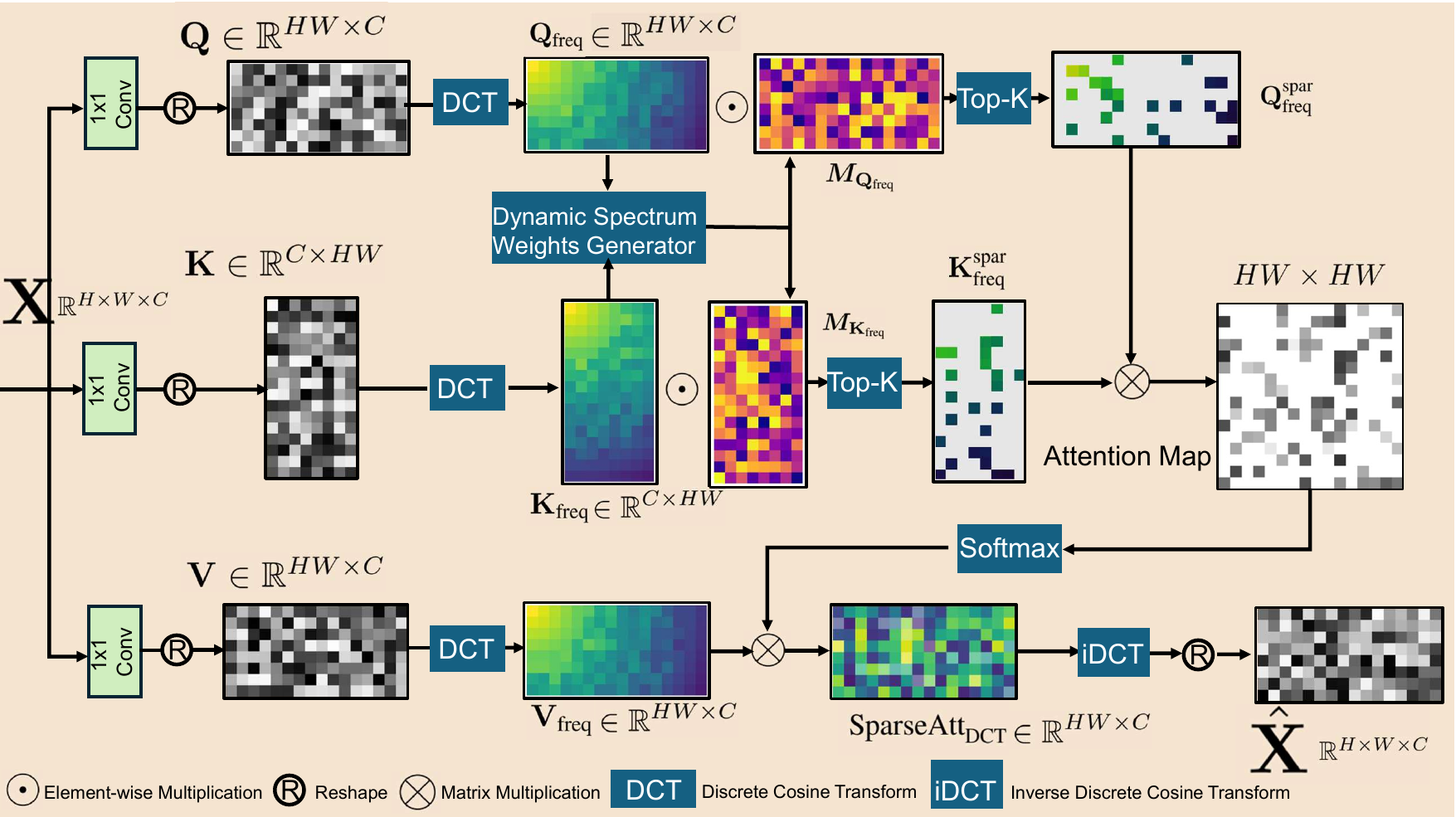}
	\caption{ Dynamic SpectraFormer Block with sparse spectrum attention. We first convert the query ($\mathbf{Q}$), key ($\mathbf{K}$), and value ($\mathbf{V}$) matrices into the frequency domain by using the Discrete Cosine Transform (DCT). Next, a Dynamic Spectrum Weight Generator (DSWG) adaptively selects $K$ frequency bands based on the input underwater image's features. Thirdly, we use the sparsified $\mathbf{Q}$ and $\mathbf{K}$ matrices to compute the attention. Finally, we use the inverse DCT (iDCT) to transform the value matrix ($\mathbf{V}$) to the spatial domain. Using sparse attention in the DCT domain, the Dynamic SpectraFormer with global attention capability could capture long-term dependency without sacrificing its universal approximation power.  
}
\end{figure*}
\begin{figure*}[htbp]
	\centering
	\includegraphics[width=0.88\linewidth]{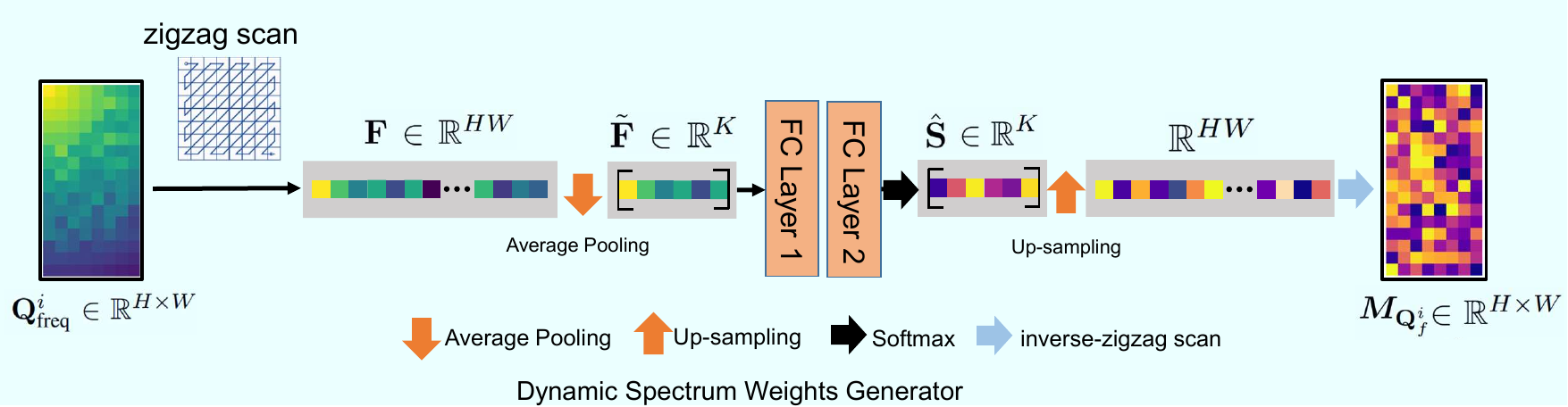}
	\caption{The Dynamic Spectrum Weights Generator. Our DSWG module aims to produce dynamic spectrum weights $\boldsymbol{M}$ to modify the frequency bands of the transformed DCT, and plays a role as a frequency band enhancer.}
\end{figure*}

Fig. 1 illustrates the DCT spectrum of an underwater image before and after enhancement. The enhanced image, with its restored details, demonstrates an increased presence of high-frequency components. This observation leads us to naturally raise a question: \textit{Can we enhance the underwater image in the frequency domain in an efficient way?}
 In this paper, we attempt to answer this question and address the issues mentioned above by proposing a Dynamic SpectraFormer, which could dynamically enhance the underwater image in the frequency domain. Our proposed method could manipulate spectrum bands concerning their contents and focus on these key frequencies, and efficiently capture the essential details necessary for image enhancement while discarding redundant information.

The core module of Dynamic SpectraFormer involves applying DCT to convert the query ($\mathbf{Q}$), key ($\mathbf{K}$), and value ($\mathbf{V}$) matrices into the frequency domain.  After computing the attention across these refined frequency components, we employ the inverse DCT (iDCT) to transform the attended feature back to the spatial domain. 
Overall, our contributions can be summarized as follows:
\begin{itemize}
	\item We developed an efficient spectrum band attention module named Dynamic SpectraFormer with global attention capability, which could balance computational efficiency and the ability to capture long-term dependencies.

 \item To adaptively amplify the useful frequency bands while downplaying others, we designed a Dynamic Spectrum Weight Generator (DSWG).  The DSWG, with a simple architecture, could empower the backbone with the ability to reweight high-frequency and low-frequency elements concerning the underwater image content. 
	\item The Dynamic SpectraFormer achieved unparalleled performance across widely recognized underwater image datasets.
\end{itemize}

\section{Related Works}
\subsection{Underwater Image Enhancement}

\textit{Non Learning-based Methods:}
In this section, we delve into the existing methodologies for enhancing underwater images, broadly classifying them into two fundamental categories: approaches that do not rely on physical models and those that are grounded in physical modeling. Model-free approaches focus on enhancing image clarity by directly adjusting pixel values, methods like multi-scale fusion \cite{Ancuti2018, Ancuti2012}, and pixel distribution modification \cite{Ghani2015}. While these techniques are straightforward, they may neglect essential underwater imaging dynamics, possibly introducing artifacts into complex aquatic environments. In contrast, model-based methods \cite{Li2016, He2010} treat enhancement as a complex inverse problem, inferring parameters through well-defined priors, red channel attenuation \cite{He2010}, and minimum information loss \cite{Li2016}. However, their effectiveness may be compromised in extreme conditions, and struggle to ensure consistent and reliable image enhancement across various scenarios.
\textit{Deep Learning-dependent Methods:} Advancements in deep learning have significantly influenced underwater image enhancement. Pritish \textit{et al.}~\cite{Pritish2019Adversarial} applied adversarial learning to improve underwater image quality. In addition, Li \textit{et al.}~\cite{Li2026WaterNet} introduced a network that utilizes a gated fusion module, integrating gamma-corrected, contrast-enhanced, and white-balanced inputs for enhanced image quality. Jiang \textit{et al.}~\cite{Jiang2020Perceptual} designed a cutting-edge perceptual adversarial network, optimized for underwater imagery, which adaptively combines latent features to mitigate image degradation. Li \textit{et al.}~\cite{Li2027WaterGAN} proposed WaterGAN, which generates synthetic underwater-style images from terrestrial scenes and depth data. Yang \textit{et al.}~\cite{Yang2020cGAN} developed a conditional generative adversarial network (cGAN) aimed at enhancing the visual clarity of underwater images.

\subsection{Vision Transformer and Frequency Domain Learning}
\textit{Vision Transformer (ViT)} \cite{dosovitskiy2020image} is the pioneering work that crops the entire image into $16\times16$ patches and treats each patch as a token as the input for the transformer. By utilizing an extremely large-scale dataset, JFT-300M \cite{sun2017revisiting}, ViT achieved promising performance comparable to CNN-based backbones.
To further lower the computational complexity, Swin Transformer\cite{liu2021swin} proposed a new local attention paradigm that employs patch-level multi-head attention equipped with a hierarchical fusion design. 
It has been demonstrated that the Fourier transforms might replace the multi-headed attention layers in transformers and produce equivalent performance. The Fourier transform-based approaches, FNet \cite{eckstein2022fnet}, GFN \cite{rao2021global}, and AFNO \cite{guibas2021adaptive}, proposed to mix the tokens by utilizing Fast Fourier Transform (FFT) in the frequency domain and achieved remarkable accuracy in visual recognition tasks. However, since the GFN uses static global filters, which are unchanged for different input images, to exploit the long-term interaction information of spectrum tokens, we argue that the image-agnostic global filter is not the optimal choice. These Fourier transform-based methods are thought to be inefficient at capturing high frequencies that primarily carry local information, as they treat all the spectrum equally.

\textit{Frequency Domain Learning}
CNN-empowered Frequency domain learning has been successfully applied in multiple vision tasks, including low-level vision such as JPEG image compression \cite{gueguen2018faster}, and image super-resolution \cite{magid2021dynamic}, as well as  high-level vision, such as frequency domain attention FcaNet \cite{qin2021fcanet}, which enhances the representability of ResNet \cite{he2016deep} on the image classification task. Specifically, the work proposed by Xu \textit{et al.} in \cite{xu2020learning} finds that the down-sampling in the frequency domain can better preserve image information than spatially resizing the images. This motivated us to derive frequency token interaction information in the down-sampled frequency domain.
\section{Approach}
In this section, we present the proposed Dynamic SpectraFormer in detail.
After briefly introducing the overall architecture as shown in Fig. 2, we present the Dynamic SpectraFormer block. Our work aims to harness the power of DCT to boost the performance of Transformer models for high-definition underwater image enhancement tasks. 

\subsection{Preliminaries: Discrete Cosine Transform}
In comparison to the Discrete Fourier Transform (DFT), the DCT is a real-valued transform that also breaks down a given signal or picture into its frequency components. Therefore, in terms of computational cost, DCT is more suited for deep neural networks.
Mathematically, the two-dimensional (2D) DCT is formatted as follows:
\begin{equation}
	B_{h, w}^{i, j}=\cos \left(\frac{\pi h}{H}\left(i+\frac{1}{2}\right)\right) \cos \left(\frac{\pi w}{W}\left(j+\frac{1}{2}\right)\right) .
\end{equation}
Then the 2D DCT for the input image with width $W$ and height $H$ is written by:
\begin{equation}
	\begin{gathered}
		f_{h, w}^{2 d}=\sum_{i=0}^{H-1} \sum_{j=0}^{W-1} I_{i, j}^{2 d} B_{h, w}^{i, j} \\
		\text { s.t. } h \in\{0,1, \cdots, H-1\}, w \in\{0,1, \cdots, W-1\},
	\end{gathered}
\end{equation}

where $f^{2 d} \in \mathbb{R}^{H \times W}$ is the 2D DCT frequency spectrum, $I^{2 d} \in \mathbb{R}^{H \times W}$ is the input image, Accordingly, the inverse 2D DCT for the image is formulated as:
\begin{equation}
	\begin{gathered}
		I_{i, j}^{2 d}=\sum_{h=0}^{H-1} \sum_{w=0}^{W-1} f_{h, w}^{2 d} B_{h, w}^{i, j} \\
		\text { s.t. } i \in\{0,1, \cdots, H-1\}, j \in\{0,1, \cdots, W-1\} .
	\end{gathered}
\end{equation}
\par The fast algorithm in \cite{makhoul1980fast} can be utilized to reduce the complexity of 1D-DCT from $\mathcal{O}\left(N^2\right)$ to $\mathcal{O}(N \log N)$.
\subsection{Revisit Scaled Dot-Product Attention}

In transformer architectures, an input sequence is represented by a set of vectors \((x_1, \ldots, x_N)\), or more formally, \(\mathbf{X} \in \mathbb{R}^{N \times C}\). Within these models, self-attention operates by mapping each vector \(x_i\) to a trio of vectors known as query (\(\mathbf{Q}\)), key (\(\mathbf{K}\)), and value (\(\mathbf{V}\)), through unique linear transformations \(\mathbf{W}^Q\), \(\mathbf{W}^K\), and \(\mathbf{W}^V\), respectively, each with a shape of \(\mathbb{R}^{C \times C}\). This transformation process yields matrices \(\mathbf{Q}\), \(\mathbf{K}\), and \(\mathbf{V}\) in \(\mathbb{R}^{N \times C}\). In the original Transformer framework, the concept of scaled dot-product attention is formulated as follows:

\begin{equation}
\label{eq:modified-att}
\text{Att}(\mathbf{X}) = \text{softmax}\left(\frac{\mathbf{Q}\mathbf{K}^T}{\sqrt{C}}\right)\mathbf{V}.
\end{equation}

For vision transformers, the input is a concatenation of vectorized feature maps \(\mathbf{X}^{\prime} \in \mathbb{R}^{H \times W \times C}\), flattened into a sequence of vectors \(x_1, \ldots, x_{HW}\), and the sequence length becomes \(N = HW\). In high-resolution scenarios, where \(N\) is substantially larger than \(C\), the computational load of the softmax operation in Eq. 4 becomes intractable, yielding a computational complexity of \(\mathcal{O}((HW)^2C)\). Although the sequence length can be curtailed and thus computational demands can be decreased with the use of local-window techniques \cite{peng2023u, ren2022reinforced}, the power of capturing long-term dependencies, is weakened. To address this issue, we introduce Dynamic SpectraFormer with global attention capability, which could capture the long-term dependency via sparse attention in the DCT domain without losing the universal approximating power.

\subsection{Sparse Spectrum Attention}

 Given the input feature-map \( \mathbf{X} \)  with dimensions \(H \times W \times C\),  our Sparse Spectrum Attention Block first generates \emph{query} (\(\mathbf{Q}\)), \emph{key} (\(\mathbf{K}\)) and \emph{value} (\(\mathbf{V}\)) projections, with encode local image features. To accomplish this, \(1 \times 1\) convolution is applied to the aggregate pixel-wise cross-channel context. This results in the following: \(\mathbf{Q} = \mathbf{W}_d^Q  \mathbf{X}\), \(\mathbf{K} = \mathbf{W}_d^K  \mathbf{X}\) and \(\mathbf{V} = \mathbf{W}_d^V  \mathbf{X}\). Where \(\mathbf{W}_d(\cdot)\) is \(1 \times 1\) point-wise convolution. 
Then, we reshape the projections into 
\( \mathbf{Q} \in \mathbb{R}^{HW \times C} \), \( \mathbf{K} \in \mathbb{R}^{C \times HW} \), and \( \mathbf{V} \in \mathbb{R}^{HW \times C} \) matrix.

After this we transform \(\mathbf{Q}\), \(\mathbf{K}\), and \(\mathbf{V}\) into the frequency domain via DCT denoted as:
\begin{align}
\mathbf{Q}_{\text{freq}} &= \text{DCT}(\mathbf{Q}) \in \mathbb{R}^{HW \times C}, \\
\mathbf{K}_{\text{freq}} &= \text{DCT}(\mathbf{K}) \in \mathbb{R}^{C \times HW}, \\
\mathbf{V}_{\text{freq}} &= \text{DCT}(\mathbf{V}) \in \mathbb{R}^{HW \times C},
\end{align}
To avoid the  \( \mathcal{O}(W^2H^2) \) level computation complexity, we employ Dynamic Spectrum Weights Generator (DSWG) to re-weight the \(\mathbf{Q}_{\text{freq}}, \mathbf{K}_{\text{freq}}\) and select top-\emph{K} (\( K \ll H \times W \)) informative spectrum bands for calculating the attention map. The sparse attention is formulated as follows.
\begin{align}
\mathbf{A}_{\text{spar}}\in \mathbb{R}^{HW \times HW} &= \frac{\mathcal{T}_K(\mathbf{Q}_{\text{freq}}) \mathcal{T}_K(\mathbf{K}_{\text{freq}})^{\top}}{\mathbf{\lambda}}, \\
\text{SparseAtt}_{\text{DCT}}(\mathbf{Q}_{\text{freq}}, \mathbf{K}_{\text{freq}}, \mathbf{V}_{\text{freq}}) &= \operatorname{softmax}(\mathbf{A}_{spar}) \mathbf{V}_{\text{freq}}.
\end{align}
where \(\mathcal{T}_K(\cdot)\) indicates the DSWG followed by top-\emph{K} operation, and it promotes sparsity and focuses attention on significant features. \( \mathbf{A}_{\text{spar}}\in \mathbb{R}^{HW \times HW} \) is the sparse attention map. The computational complexity for the sparse attention module is $\mathcal{O}(HWKC)$, where $K \ll HW$ is the number of key frequency bands that have been chosen.  In contrast, pixel-level attention has complexity of $\mathcal{O}(H^2W^2C)$. 
 \\After Sparse Spectrum Attention in the frequency domain, we transform the enhanced features back into the spatial domain. This step is performed via the Inverse Discrete Cosine Transform (IDCT):
\begin{equation}
\hat{\mathbf{X}} = \text{IDCT}\left(\text{SparseAtt}_{\text{DCT}}(\mathbf{Q}_{\text{freq}}, \mathbf{K}_{\text{freq}}, \mathbf{V}_{\text{freq}})\right),
\end{equation}
where \(\hat{\mathbf{X}}\) is the enhanced spacial features. 
Finally, the outputs across all attention heads are concatenated and transformed via a linear normalization layer. The architecture of the Dynamic SpectraFormer block is detailed in Fig. 3. Note that the linear normalization layer is not drawn in the figure.

\begin{table*}[ht!]
\centering
\caption{Quantitative comparison on the UIEB, LSUI, and EUVP underwater datasets. Note that the FLOPs is calculated with image size $256\times256$ pixels. The best results are shown in \textbf{bold} text.}
\label{tab:merged}
\begin{tabular}{
    l
    S[table-format=2.2]| 
    S[table-format=2.2]|
    S[table-format=2.2]
    S[table-format=1.4]|
    S[table-format=1.4]|
    S[table-format=1.4]
    S[table-format=3.0] 
    S[table-format=3.2] 
}
\toprule
Method & \multicolumn{3}{c}{PSNR} & \multicolumn{3}{c}{SSIM} & \multicolumn{1}{c}{Params} & \multicolumn{1}{c}{FLOPs} \\
& \multicolumn{3}{c}{(UIEB | LSUI | EUVP)} & \multicolumn{3}{c}{(UIEB | LSUI | EUVP)} & \multicolumn{1}{c}{(M)} & \multicolumn{1}{c}{(GFLOPs)} \\
\midrule
Ucolor~\cite{li2021underwater} & 20.78 & 22.91 & \multicolumn{1}{c|}{-} & 0.8713 & 0.8902 & \multicolumn{1}{c|}{-} & \multicolumn{1}{r|}{157M} & 443.85G \\
WaterNet~\cite{li2019underwater} & 19.81 & 17.73 & \multicolumn{1}{c|}{20.14} & 0.8612 & 0.8223 & \multicolumn{1}{c|}{0.68} & \multicolumn{1}{r|}{25M} & 193.7G \\
UGAN~\cite{fabbri2018enhancing} & 20.68 & 19.79 & \multicolumn{1}{c|}{23.49} & 0.8430 & 0.7843 & \multicolumn{1}{c|}{0.7802} & \multicolumn{1}{r|}{57M} & 38.97G \\
FUnIE-GAN~\cite{islam2020fast} & 19.45 & 19.37 & \multicolumn{1}{c|}{23.40} & 0.8602 & 0.8401 & \multicolumn{1}{c|}{0.8420} & \multicolumn{1}{r|}{7M} & 10.23G \\
Deep SESR~\cite{islam2020simultaneous} & \multicolumn{1}{c|}{-} & \multicolumn{1}{c|}{-} & \multicolumn{1}{c|}{24.21} & \multicolumn{1}{c|}{-} & \multicolumn{1}{c|}{-} & \multicolumn{1}{c|}{0.8401} & \multicolumn{1}{r|}{3M} & 29.32G \\
U-Shape Transformer~\cite{peng2023u} & 22.91 & 24.16 & \multicolumn{1}{c|}{-} & 0.9100 & 0.9322 & \multicolumn{1}{c|}{-} & \multicolumn{1}{r|}{65.6M} & 70.2G \\ 
UHD Underwater Enhancement~\cite{wei2022uhd} & 25.04 & \multicolumn{1}{c|}{-} & \multicolumn{1}{c|}{-} & 0.9158 & \multicolumn{1}{c|}{-} & \multicolumn{1}{c|}{-} & \multicolumn{1}{r|}{157M} & 70.23G \\
\textbf{Ours} & \textbf{25.98} & \textbf{26.33} & \multicolumn{1}{c|}{\textbf{29.78}} & \textbf{0.9341} & \textbf{0.9327} & \multicolumn{1}{c|}{\textbf{0.8848}} & \multicolumn{1}{r|}{64.2M} & 50.2G \\
\bottomrule
\end{tabular}
\end{table*}


\subsection{Dynamic Spectrum Weights Generator (DSWG)}
According to the property of DCT, a single output element from DCT has a component of each of the input pixels in the spatial domain. To this end, the goal of our DSWG module is to generate dynamic spectrum weights $\boldsymbol{M}$ to modulate the transformed DCT frequency bands and act as a frequency band enhancer. The process of dynamic spectrum weight generation regarding the Query in frequency $\mathbf{Q}^{i}_{\text{freq}} \in \mathbb{R}^{H \times W}$, where $i$ is the channel index, is shown in Fig. 4. Because the spectrum bands of DCT have clear physical meaning and are arranged from left to right and top to bottom in a strictly increasing order of frequencies, we can flatten all the spectrum in $\mathbf{Q}^{i}_{\text{freq}} \in \mathbb{R}^{H \times W }$ in to $\mathbf{F}\in \mathbb{R}^{HW }$ by using a zigzag scan, and obtain the one-dimensional embedding. 

\par Intuitively, we can employ multiple fully connected layers to capture the full spectrum band interaction information. However, the computation complexity will be raised significantly to $\mathcal{O}(H^2 W^2 )$. Sparsification is one approach to solve this problem, as we argue that not all information in the frequency range contributes to perceptual ability of the transformer. Many DCT-based methods introduce sparsity to DCT blocks through quantization \cite{liu2022nommer}. Following this, we propose to only utilize the average-pooling operation over the spectrum embedding $\mathbf{F}$. We down-sample the frequency band $\mathbf{F} \in \mathbb{R}^{HW}$ to $\tilde{\mathbf{F}} \in \mathbb{R}^{K}$. Then, a two-layer FC design is used to obtain the spectrum band attention weight. The process addressed above can be formulated as follows:
\begin{equation}
	\tilde{\mathbf{F}} \in \mathbb{R}^{K}= \mathbf{DownSampling}(\mathbf{zigzag}(\mathbf{Q}^{i}_{\text{freq}})),
\end{equation}
\begin{equation}
	\mathbf{S} \in \mathbb{R}^{K} =\textrm{reshape}(\mathbf{W}_2 \sigma\left[\mathbf{W}_1 \operatorname{LayerNorm}(\tilde{\mathbf{F}})\right])
\end{equation}

where, $\mathbf{DownSampling}$ and $\mathbf{zigzag}$ are average-pooling and zigzag scan operation, respectively. $\sigma$ is the activation function implemented by Gaussian Error Linear Units (GELU) \cite{hendrycks2016gaussian}, LayerNorm $(\cdot)$ represents the layer normalization \cite{ba2016layer}. 
$\mathbf{W}_1 \in \mathbb{R}^{K \times V}$ denote the weights of a fully-connected layer, increasing the feature dimension from $K$ to $V$ where $V$ is a fixed value. $\mathbf{W}_2 \in \mathbb{R}^{V \times K }$ refers to the weights of a fully-connected layer reshaping the feature from $V$ back to the original dimension $K$. 
\par Finally, the $\mathbf{S}$ is further processed by $\textrm{softmax}$ function and followed by an inverse zigzag scanning mapping $\mathbb{R}^{HW} \mapsto \mathbb{R}^{H \times W}$ that reshape the embedding $\hat{\mathbf{F}}$ back to get the dynamic spectrum weights $\boldsymbol{M}_{\mathbf{Q}_{freq}^{i}}$ as follows:
\begin{equation}
	\hat{\mathbf{S}} = \textrm{softmax}(\mathbf{S})
\end{equation}
\begin{equation}
	\boldsymbol{M}_{\mathbf{Q}_{freq}^{i}} = \mathbf{inverse\text{-}zigzag}(\mathbf{UpSampling}(\hat{\mathbf{S}}))
\end{equation}
$\mathbf{UpSampling}$ and $\mathbf{inverse-zigzag}$ are up-pooling and inverse zigzag scan operation, respectively.

\subsection{In-Depth View of Dynamic SpectraFormer's Design}

 The Dynamic SpectraFormer's overall architecture is illustrated in Fig. 2, incorporating an encoder-decoder structure. For a given low-quality image, denoted as $\mathbf{I}_{\text{degraded}} \in \mathbb{R}^{H \times W \times 3}$, our approach initiates with $3\times3$ convolutions to extract the enriched low-level feature. Next, these low-level features go through a 4-level symmetric encoder-decoder. 
The encoder-decoder network is equipped with a sequence of Dynamic SpectraFormer blocks. The encoder increases channel numbers while hierarchically reducing spatial dimension. The decoder gradually recovers the high-resolution representations from the low-resolution latent features $\mathbf{D}$~$\in$~$\mathbb{R}^{\frac{H}{8}\times\frac{W}{8}\times 8C}$. 
We use the pixel-unshuffle and pixel-shuffle operations~\cite{shi2016real} for feature downsampling and upsampling, respectively. 
The encoder features and the decoder features are concatenated via skip connections~\cite{ronneberger2015u} to aid the recovery procedure. 
In the final step, a convolutional layer processes these polished features to construct the final restored image  $\mathbf{R}$~$\in$~$\mathbb{R}^{H\times W \times 3}$.
\begin{figure*}[htbp]
	\centering
	\includegraphics[width=0.80\linewidth]{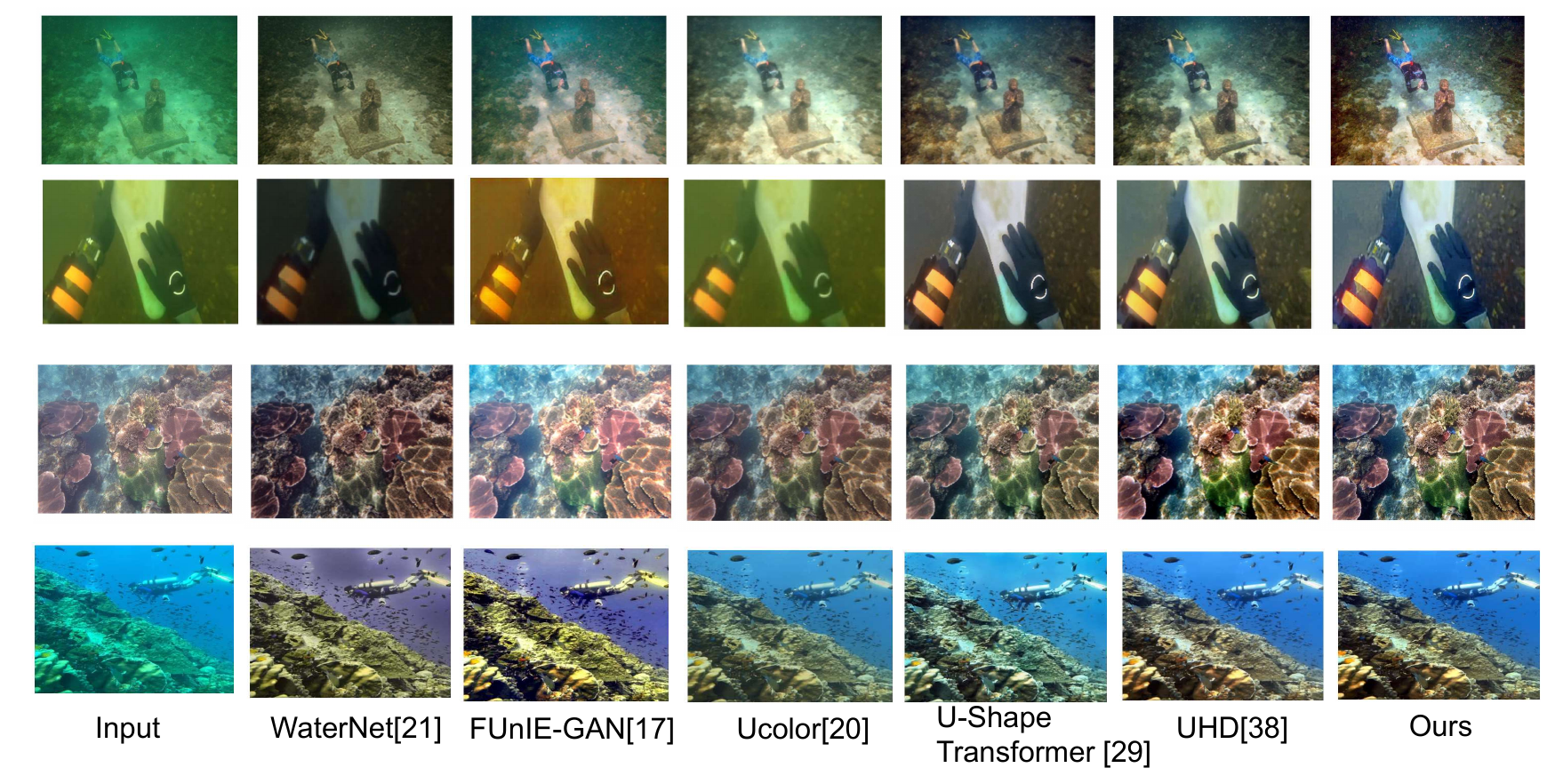}
	\caption{Visual comparison on the underwater dataset. }
  \vspace{-5mm}
\end{figure*}

We utilize a hybrid loss function, which balances pixel accuracy and structural integrity to guide the network to enhance the degraded image:
\begin{equation}
    \mathcal{L_{\text{total}}} = w_1\mathcal{L_{\text{pixel}}}  + w_2 \mathcal{L_{\text{MS-SSIM}}},
\end{equation}
\begin{equation}
\mathcal{L_{\text{pixel}}} = \left\| I_{\text{degraded}} - I_{\text{gt}} \right\|_1
\end{equation}
where \( I_{\text{gt}} \) is the ground truth, and the L1 norm \( \left\| \cdot \right\|_1 \) aims to reduce the absolute differences between the degraded image and the true image. $\mathcal{L_{\text{MS-SSIM}}}$ represents the Multiscale Structural Similarity Index (MS-SSIM)\cite{wang2003multiscale}. \( w_1 \) and \( w_2 \) are weights that were empirically set through experimental validation to ensure an optimal trade-off between global coherence and local texture fidelity. In our evaluation, the \( w_1 \) and \( w_2 \) were set to be 0.6 and 0.4, respectively.

\subsection{Complexity Analysis}
The sparse attention module exhibits a computational complexity of $\mathcal{O}(HWKC)$, where $K \ll HW$ denotes the number of selected key frequency bands, significantly reduced from full pixel-level attention complexity: $\mathcal{O}(H^2W^2C)$. 

For the DCT and inverse DCT layers, applied per channel, the complexity is dictated by the fast DCT algorithm, yielding a complexity of $\mathcal{O}(H W C \lceil \log_2(H W) \rceil)$ for both forward and inverse transformations.

The DSWG module, given the input size represented by $\boldsymbol{M}_{\mathbf{Q}_{\text{freq}}^{i}}$ and weight matrices $\mathbf{W}_1 \in \mathbb{R}^{K \times V}$ and $\mathbf{W}_2 \in \mathbb{R}^{V \times K}$, the DSWG's computational complexity is calculated as $\mathcal{O}(K V^2 C + K^2 V C)$. To this end, the overall complexity is: $\mathcal{O}(HWKC + H W C \lceil \log_2(H W) \rceil + K V^2 C + K^2 V C)$. 
See Fig. 7 for detailed comparison results.

\begin{figure}[htbp]
	\centering
	\includegraphics[width=0.92\linewidth]{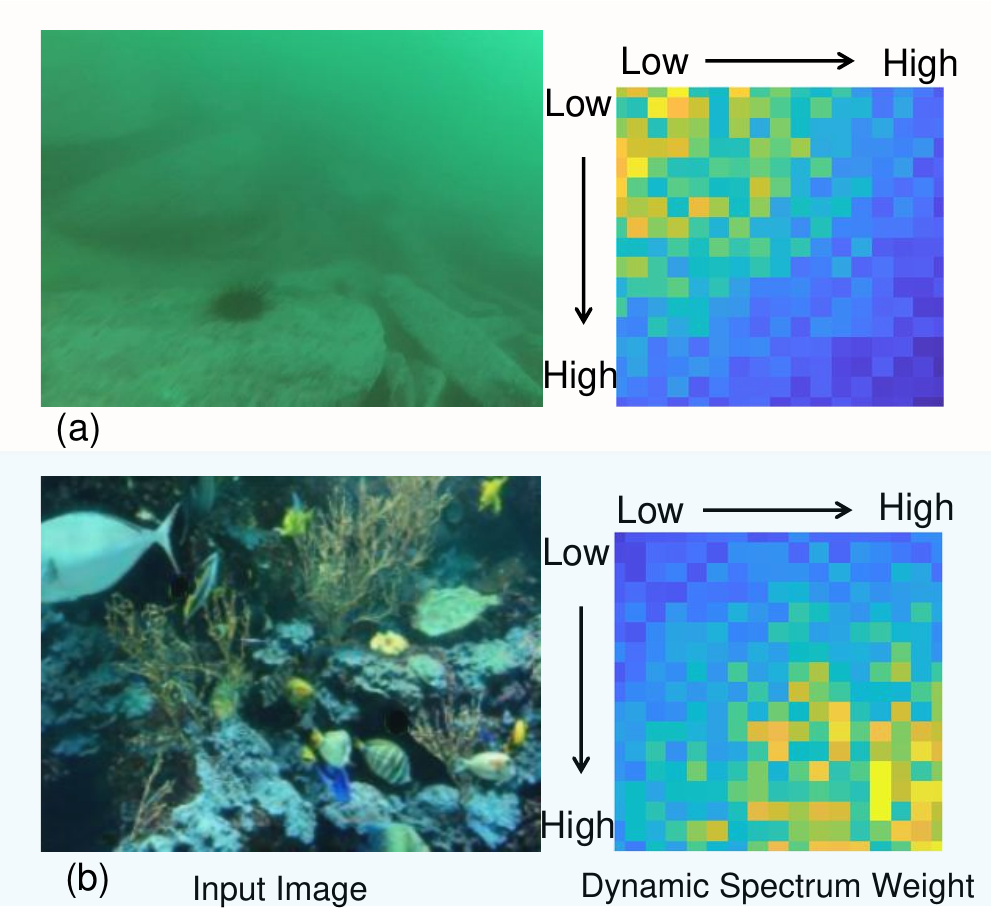}
	\caption{ The images from LSUI data set \cite{peng2023u} with corresponding dynamic spectrum weights. For the images containing rich details, as shown in (b), the generated dynamic spectrum weight mostly focuses on the high-frequency bands, while for the plain images (a), dynamic spectrum weight emphasizes the low-frequency bands on the contrary. To this end, our Dynamic Spectrum Weights Generator could adaptively reweight the high-frequency and low-frequency components concerning the image content and facilitate the token mixing operation. Note that the lighter the color, the larger the spectrum weight. }
\end{figure}

\section{Experimental Results}
\subsection{Experimental Setting}

\begin{table}[tbp]
\centering
\caption{Effectiveness of the Dynamic Spectrum Weights Generator in Dynamic SpectraFormer on three datasets. }
\label{table:module_effectiveness}
\resizebox{\columnwidth}{!}{%
\begin{tabular}{@{}lcccccc@{}}
\toprule
Modules & \multicolumn{2}{c}{UIEB} & \multicolumn{2}{c}{LSUI} & \multicolumn{2}{c}{EUVP} \\
 & PSNR & SSIM & PSNR & SSIM & PSNR & SSIM \\
\midrule
Random spectrum band & 21.01 & 0.8402 & 24.19 & 0.8105 & 27.20 & 0.8321 \\
All-pass filter & 23.72 & 0.8461 & 25.20 & 0.8753 & 28.09 & 0.8654 \\
\textbf{Dynamic Spectrum Weight Generator} & \textbf{25.98} & \textbf{0.9341} & \textbf{26.33} & \textbf{0.9327} & \textbf{29.78} & \textbf{0.8848} \\
\bottomrule
\end{tabular}%
}
\end{table}

\begin{table}[tbp]
\centering
\caption{Effectiveness of the Sparsity of Sparse Spectrum Attention.}
\label{table:loss_function_effectiveness}
\resizebox{\columnwidth}{!}{%
\begin{tabular}{@{}lcccccc@{}}
\toprule
Spectrum length (top-\emph{K}) & \multicolumn{2}{c}{UIEB} & \multicolumn{2}{c}{LSUI} & \multicolumn{2}{c}{EUVP} \\
 & PSNR & SSIM & PSNR & SSIM & PSNR & SSIM \\
\midrule
8 & 19.45 & 0.8657 & 19.71 & 0.7812 & 19.32 & 0.7782 \\
16 & 22.19 & 0.9038 & 23.74 & 0.8462 & 26.22 & 0.8578 \\
32 & 24.23 & 0.9212 & 24.82 & 0.8570 & 28.01 & 0.8721 \\
64 & 25.98 & \textbf{0.9341} & \textbf{26.33} & \textbf{0.9327} & \textbf{29.78} & \textbf{0.8848} \\
128 & \textbf{26.00} & 0.9340 & 26.33 & 0.9329 & 29.82 & 0.8810 \\
256 & 25.97 & \textbf{0.9341} & 26.19 & 0.9218 & 29.64 & 0.8801 \\
\bottomrule
\end{tabular}%
}
\end{table}
To facilitate a comprehensive comparative analysis, we integrate three underwater datasets for our experimental evaluation, described as follows:

For the dataset preparation, the setting approach in \cite{peng2023u} was employed. The LSUI \cite{peng2023u} dataset was segmented into 4500, and 404 images for training and testing, respectively.  In the evaluation stage, we also conducted assessments using the underwater image sets from UIEB (90 pairs)\cite{li2019underwater}, LSUI (504 pairs)\cite{peng2023u}, and EUVP (515 pairs)\cite{islam2020fast}, respectively. Evaluation metrics include Peak Signal Noise Ratio (PSNR) and Structural Similarity Index (SSIM), measuring the color and structural fidelity between enhanced images and ground truths.
\subsection{Training Process Details}
For the generation of images in our training set, we employed data augmentation techniques including horizontal and vertical flipping, noise addition, and contrast variation. All input images were resized to dimensions of $512\times512$ pixels for consistency. During the training process, we utilized the Adam optimizer \cite{kingma2014adam} with an initial learning rate of $3 \times 10^{-4}$, adjusting it via the cosine annealing strategy. Our network was implemented using PyTorch and trained on 4 $\times$ NVIDIA Tesla A100s.

\subsection{Comparisons with State-of-the-Art Methods}
\textbf{Quantitative Evaluation}
To consolidate our performance superiority, we compare our SpectraFormer with multiple UIE methods, including Ucolor~\cite{li2021underwater}, WaterNet~\cite{li2019underwater}, UGAN~\cite{fabbri2018enhancing}, FUnIE-GAN~\cite{islam2020fast}, Deep SESR~\cite{islam2020simultaneous}, U-Shape Transformer~\cite{peng2023u}, and UHD Underwater image enhancement ~\cite{wei2022uhd}.
The experimental results are detailed in Tab. I and demonstrate a significant numerical superiority of our approach over most current techniques. In particular, our method demonstrates the best performance, achieving a 0.94 $dB$ improvement in PSNR on the UIEB benchmark compared to the recently proposed UHD Underwater Enhancement~\cite{wei2022uhd}. It is noteworthy that our proposed Dynamic SpectraFormer achieves the best performance for all three databases. Furthermore, we achieve an increment of 3.07 $dB$ in UIEB over the latest transformer-based approach Ushape~\cite{peng2023u} with fewer FLOPs (50.2 GFLOPs vs. 70.2 GFLOPs). 

\textbf{Qualitative Evaluation}
Fig. 5 illustrates the results of our method applied to UHD image enhancement for the UIEB dataset, along with results from other approaches. We notice that some conventional methods, such as Ucolor~\cite{li2021underwater}, tend to overdo the enhancement and lead to color distortions. Methods based on GANs, as seen in UGAN~\cite{fabbri2018enhancing} and FUnIE-GAN~\cite{islam2020fast}, struggle to revive the full-color spectrum and are prone to introducing unnatural patterns. Even recent deep-learning solutions such as U-Shape Transformer~\cite{peng2023u} display issues like mixed-up details and inaccurate hues due to their limited ability to capture complex image features. Furthermore, UHD UIE relies on an extensive network of convolution layers to improve performance as introduced in ~\cite{wei2022uhd}, making them less suited for handling UHD images on devices (e.g. AUVs) with limited computing power.
\begin{figure}[tbp]
	\centering
	\includegraphics[width=1\linewidth]{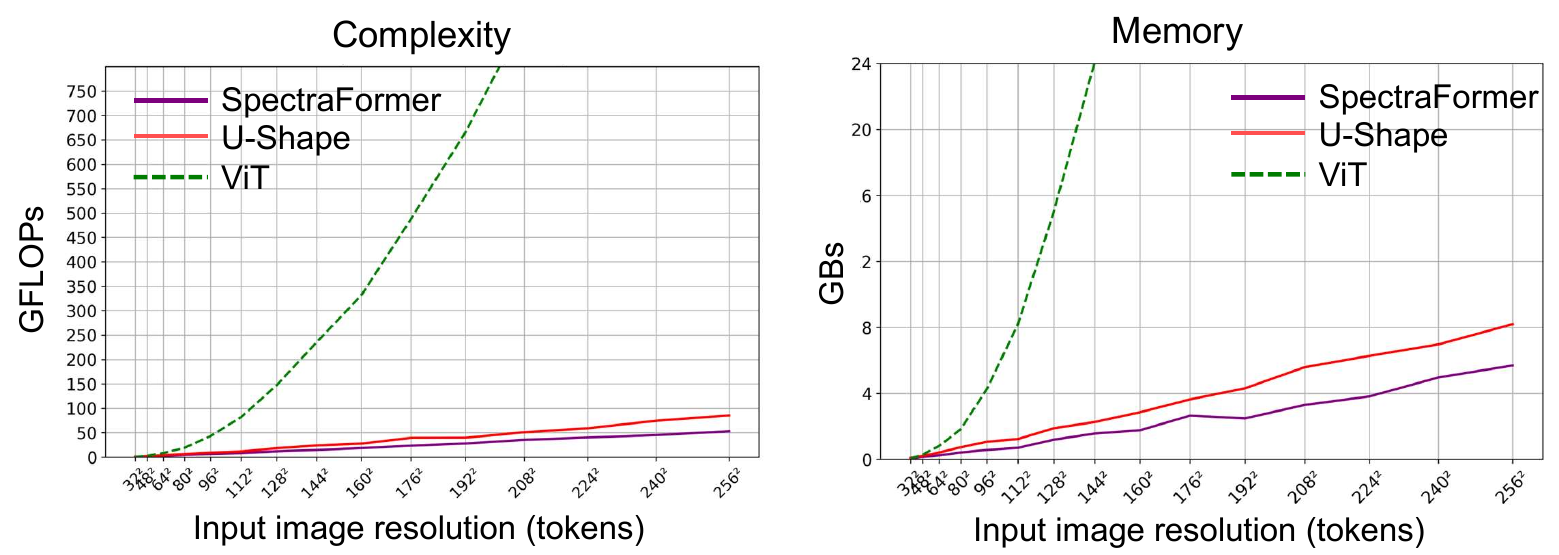}
	\caption{ Comparisons among ViT \cite{dosovitskiy2020image}, U-Shape Transformer~\cite{peng2023u} and our Dynamic SpectraFormer in (a) FLOPs (b) memory consumption concerning different image resolutions. The time and memory complexity of the ViT increase quadratically with the spatial resolution of input \( \mathcal{O}(W^2H^2) \), while our method significantly lowers the computational complexity, as shown in the figure.}
\end{figure}
In contrast, our approach demonstrates proficiency in processing UHD images, effectively restoring synthetic colors and clear definitions. As depicted in Fig. 5, images enhanced by our method exhibit a fidelity that is much closer to their true, unblemished state, underlining the practical effectiveness of our method.

\subsection{Ablation Studies and Analysis}
To demonstrate the efficacy of the proposed components, we undertake the subsequent ablation studies on the UIEB dataset \cite{li2019underwater}.
\subsubsection{Effectiveness of the Dynamic Spectrum Weights Generator in Dynamic SpectraFormer}


To evaluate the effectiveness of the proposed DSWG module, we designed three versions of experiments. (1) Randomly select the spectrum bands after DCT for all $\mathbf{Q}$, $\mathbf{K}$, and $\mathbf{V}$. (2) All-pass filter, i.e., set all the weights to 1, leaving the spectrum untouched. In other words, in the frequency domain representation, the frequency components in $X_{\text{freq}}$ are sorted according to the energy of each frequency component (i.e., the absolute value of the DCT coefficients), and then the top $K$ frequency components with the highest energy are selected. (3) Our DSWG with top-\emph{K}. The comparison results are listed in Tab. II. We can see that the random weight leads to significant performance degradation (UIEB PSNR drops from 25.98 to 21.01 $dB$), while the all-pass filter achieves much better accuracy than the random one. Among all the weight generation mechanisms, our DSWG demonstrates the best performance. Fig. 6 shows instances of the Dynamic Spectrum Weight for the LSUI test set. This result verifies that the dynamically generated spectrum weight empowers the transformer with improved representation ability. 
\subsubsection{Effectiveness of the Sparsity of Sparse Spectrum Attention}
To evaluate the influence of spectrum length (top-\emph{K}) $K$ on the performance, we compare the performance of several settings in Tab. III. We set a customized interval range for $K$ to avoid an exhaustive search.
Intuitively, the longer spectrum length is beneficial to
fine-grained modeling details of the input tokens and tends to achieve higher image quality. 
Surprisingly, as shown in Tab. III, even employing only $32$ spectrum bands can achieve 24.23 dB in UIEB. As the length grows from $64$ to $256$, the accuracy almost remains unchanged. So we finally set the $K$ to $64$ in our experiments, which achieved the best performance and complexity trade-off. This result confirms that compressed DCT spectrums can reduce the computation cost for the proposed Dynamic SpectraFormer without sacrificing accuracy. 

\section{Conclusion}
In this paper, we proposed Dynamic SpectraFormer, a highly efficient frequency domain transformer that could enhance the UHD underwater image in the frequency domain and empower it with adaptability to the contents. Extensive analysis conducted on real-world datasets, along with comprehensive ablation studies, validates the efficacy of the proposed method for UHD underwater image enhancement tasks. Furthermore, the Dynamic SpectraFormer has applicability to other widely used image restoration tasks, such as motion deblurring, image de-rain, etc.


	{\small
	\bibliographystyle{ieee_fullname}
	\bibliography{egbib}}

@String(CVPR= {IEEE Conf. Comput. Vis. Pattern Recog.})

@String(ACCV  = {ACCV})

@String(CVPR  = {CVPR})

@article{eckstein2022fnet,
  title={FNet: Mixing Tokens with Fourier Transforms},
  author={Eckstein, Ilya and Lee-Thorp, James Patrick and Ainslie, Joshua and Ontanon, Santiago},
  year={2022}
}

@article{guibas2021adaptive,
  title={Adaptive fourier neural operators: Efficient token mixers for transformers},
  author={Guibas, John and Mardani, Morteza and Li, Zongyi and Tao, Andrew and Anandkumar, Anima and Catanzaro, Bryan},
  journal={arXiv preprint arXiv:2111.13587},
  year={2021}
}

@article{rao2021global,
  title={Global filter networks for image classification},
  author={Rao, Yongming and Zhao, Wenliang and Zhu, Zheng and Lu, Jiwen and Zhou, Jie},
  journal={Advances in Neural Information Processing Systems},
  volume={34},
  pages={980--993},
  year={2021}
}

@article{dosovitskiy2020image,
  title={An image is worth 16x16 words: Transformers for image recognition at scale},
  author={Dosovitskiy, Alexey and Beyer, Lucas and Kolesnikov, Alexander and Weissenborn, Dirk and Zhai, Xiaohua and Unterthiner, Thomas and Dehghani, Mostafa and Minderer, Matthias and Heigold, Georg and Gelly, Sylvain and others},
  journal={arXiv preprint arXiv:2010.11929},
  year={2020}
}

@inproceedings{sun2017revisiting,
  title={Revisiting unreasonable effectiveness of data in deep learning era},
  author={Sun, Chen and Shrivastava, Abhinav and Singh, Saurabh and Gupta, Abhinav},
  booktitle={Proceedings of the IEEE international conference on computer vision},
  pages={843--852},
  year={2017}
}

@inproceedings{liu2021swin,
  title={Swin transformer: Hierarchical vision transformer using shifted windows},
  author={Liu, Ze and Lin, Yutong and Cao, Yue and Hu, Han and Wei, Yixuan and Zhang, Zheng and Lin, Stephen and Guo, Baining},
  booktitle={Proceedings of the IEEE/CVF International Conference on Computer Vision},
  pages={10012--10022},
  year={2021}
}

@inproceedings{xu2020learning,
  title={Learning in the frequency domain},
  author={Xu, Kai and Qin, Minghai and Sun, Fei and Wang, Yuhao and Chen, Yen-Kuang and Ren, Fengbo},
  booktitle={Proceedings of the IEEE/CVF Conference on Computer Vision and Pattern Recognition},
  pages={1740--1749},
  year={2020}
}

@article{gueguen2018faster,
  title={Faster neural networks straight from jpeg},
  author={Gueguen, Lionel and Sergeev, Alex and Kadlec, Ben and Liu, Rosanne and Yosinski, Jason},
  journal={Advances in Neural Information Processing Systems},
  volume={31},
  year={2018}
}

@inproceedings{magid2021dynamic,
  title={Dynamic high-pass filtering and multi-spectral attention for image super-resolution},
  author={Magid, Salma Abdel and Zhang, Yulun and Wei, Donglai and Jang, Won-Dong and Lin, Zudi and Fu, Yun and Pfister, Hanspeter},
  booktitle={Proceedings of the IEEE/CVF International Conference on Computer Vision},
  pages={4288--4297},
  year={2021}
}

@inproceedings{qin2021fcanet,
  title={Fcanet: Frequency channel attention networks},
  author={Qin, Zequn and Zhang, Pengyi and Wu, Fei and Li, Xi},
  booktitle={Proceedings of the IEEE/CVF international conference on computer vision},
  pages={783--792},
  year={2021}
}

@inproceedings{he2016deep,
  title={Deep residual learning for image recognition},
  author={He, Kaiming and Zhang, Xiangyu and Ren, Shaoqing and Sun, Jian},
  booktitle={Proceedings of the IEEE conference on computer vision and pattern recognition},
  pages={770--778},
  year={2016}
}

@article{makhoul1980fast,
  title={A fast cosine transform in one and two dimensions},
  author={Makhoul, John},
  journal={IEEE Transactions on Acoustics, Speech, and Signal Processing},
  volume={28},
  number={1},
  pages={27--34},
  year={1980},
  publisher={IEEE}
}

@inproceedings{liu2022nommer,
  title={NomMer: Nominate Synergistic Context in Vision Transformer for Visual Recognition},
  author={Liu, Hao and Jiang, Xinghua and Li, Xin and Bao, Zhimin and Jiang, Deqiang and Ren, Bo},
  booktitle={Proceedings of the IEEE/CVF Conference on Computer Vision and Pattern Recognition},
  pages={12073--12082},
  year={2022}
}

@article{hendrycks2016gaussian,
  title={Gaussian error linear units (gelus)},
  author={Hendrycks, Dan and Gimpel, Kevin},
  journal={arXiv preprint arXiv:1606.08415},
  year={2016}
}

@article{ba2016layer,
  title={Layer normalization},
  author={Ba, Jimmy Lei and Kiros, Jamie Ryan and Hinton, Geoffrey E},
  journal={arXiv preprint arXiv:1607.06450},
  year={2016}
}

@inproceedings{Pritish2019Adversarial,
  title={All-in-one underwater image enhancement using domain-adversarial learning},
  author={Pritish M. Uplavikar and Zhenyu Wu and Zhangyang Wang},
  booktitle={CVPR Workshops},
  year={2019},
}

@article{Li2026WaterNet,
  title={WaterNet: A gated fusion network for underwater image enhancement},
  author={Li, Chongyi and Guo, Chunle and Ren, Wenqi and Cong, Runmin and Hou, Junhui and Kwong, Sam and Tao, Dacheng},
  journal={IEEE Transactions on Image Processing},
  volume={29},
  pages={4376--4389},
  year={2019},
}

@article{Jiang2020Perceptual,
  title={Target oriented perceptual adversarial fusion network for underwater image enhancement},
  author={Jiang, Zhiying and Li, Zhuoxiao and Yang, Shuzhou and Fan, Xin and Liu, Risheng},
  journal={IEEE Transactions on Circuits and Systems for Video Technology},
  volume={32},
  number={10},
  pages={6584--6598},
  year={2022},
}

@article{Li2027WaterGAN,
  title={WaterGAN: Unsupervised generative network to enable real-time color correction of monocular underwater images},
  author={Li, Jie and Skinner, Katherine A. and Eustice, Ryan M. and Johnson-Roberson, Matthew},
  journal={IEEE Robotics and Automation Letters},
  volume={3},
  number={1},
  pages={387--394},
  year={2017},
}

@article{Yang2020cGAN,
  title={Underwater image enhancement based on conditional generative adversarial network},
  author={Yang, Miao and Hu, Ke and Du, Yixiang and Wei, Zhiqiang and Sheng, Zhibin and Hu, Jintong},
  journal={Signal Processing: Image Communication},
  volume={81},
  pages={115723},
  year={2020},
}

@article{Ancuti2018,
  title={Color balance and fusion for underwater image enhancement},
  author={Codruta O. Ancuti and Cosmin Ancuti and Christophe De Vleeschouwer and Philippe Bekaert},
  journal={IEEE Transactions on Image Processing},
  volume={27},
  number={1},
  pages={379--393},
  year={2018}
}

@inproceedings{Ancuti2012,
  title={Enhancing underwater images and videos by fusion},
  author={Cosmin Ancuti and Codruta Orniana Ancuti and Tom Haber and Philippe Bekaert},
  booktitle={2012 IEEE Conference on Computer Vision and Pattern Recognition},
  pages={81--88},
  year={2012}
}

@article{Ghani2015,
  title={Underwater image quality enhancement through integrated color model with rayleigh distribution},
  author={Ahmad Shahrizan Abdul Ghani and Nor Ashidi Mat Isa},
  journal={Applied soft computing},
  volume={27},
  pages={219--230},
  year={2015}
}

@article{Li2016,
  title={Underwater image enhancement by dehazing with minimum information loss and histogram distribution prior},
  author={Chong-Yi Li and Ji-Chang Guo and Run-Min Cong and Yan-Wei Pang and Bo Wang},
  journal={IEEE Transactions on Image Processing},
  volume={25},
  number={12},
  pages={5664--5677},
  year={2016}
}

@article{He2010,
  title={Single image haze removal using dark channel prior},
  author={Kaiming He and Jian Sun and Xiaoou Tang},
  journal={IEEE transactions on pattern analysis and machine intelligence},
  volume={33},
  number={12},
  pages={2341--2353},
  year={2010}
}

@inproceedings{li2019underwater,
title={An underwater image enhancement benchmark dataset and beyond},
author={Li, Chongyi and Guo, Chunle and Ren, Wenqi and Cong, Runmin and Hou, Junhui and Kwong, Sam and Tao, Dacheng},
booktitle={IEEE Transactions on Image Processing},
volume={29},
pages={4376--4389},
year={2019},
organization={IEEE}
}

@inproceedings{shi2016real,
title={Real-time single image and video super-resolution using an efficient sub-pixel convolutional neural network},
author={Shi, Wenzhe and Caballero, Jose and Huszar, Ferenc and Totz, Johannes and Aitken, Andrew P and Bishop, Rob and Rueckert, Daniel and Wang, Zehan},
booktitle={Proceedings of the IEEE conference on computer vision and pattern recognition},
pages={1874--1883},
year={2016}
}

@article{islam2020fast,
  title={Fast underwater image enhancement for improved visual perception},
  author={Islam, Md Jahidul and Xia, Youya and Sattar, Junaed},
  journal={IEEE Robotics and Automation Letters},
  volume={5},
  number={2},
  pages={3227--3234},
  year={2020}
}

@article{islam2020simultaneous,
  title={Simultaneous enhancement and super-resolution of underwater imagery for improved visual perception},
  author={Islam, M.J. and Luo, P. and Sattar, J.},
  journal={arXiv preprint arXiv:2002.01155},
  year={2020}
}

@article{peng2023u,
  title={U-shape transformer for underwater image enhancement},
  author={Peng, Lintao and Zhu, Chunli and Bian, Liheng},
  journal={IEEE Transactions on Image Processing},
  year={2023},
  publisher={IEEE}
}

@inproceedings{fabbri2018enhancing,
  title={Enhancing underwater imagery using generative adversarial networks},
  author={Fabbri, C. and Islam, M. J. and Sattar, J.},
  booktitle={Proceedings of the IEEE International Conference on Robotics and Automation},
  pages={7159--7165},
  year={2018},
  organization={IEEE}
}

@article{li2021underwater,
  title={Underwater image enhancement via medium transmission-guided multi-color space embedding},
  author={Li, Chongyi and Anwar, Saeed and Hou, Junhui and Cong, Runmin and Guo, Chunle and Ren, Wenqi},
  journal={IEEE Transactions on Image Processing},
  volume={30},
  pages={4985--5000},
  year={2021},
  publisher={IEEE}
}

@InProceedings{Khan_2024_WACV,
    author    = {Khan, Raqib and Mishra, Priyanka and Mehta, Nancy and Phutke, Shruti S. and Vipparthi, Santosh Kumar and Nandi, Sukumar and Murala, Subrahmanyam},
    title     = {Spectroformer: Multi-Domain Query Cascaded Transformer Network for Underwater Image Enhancement},
    booktitle = {Proceedings of the IEEE/CVF Winter Conference on Applications of Computer Vision (WACV)},
    month     = {January},
    year      = {2024},
    pages     = {1454-1463}
}

@article{sharma2023wavelength,
  title={Wavelength-based attributed deep neural network for underwater image restoration},
  author={Sharma, Prasen and Bisht, Ira and Sur, Arijit},
  journal={ACM Transactions on Multimedia Computing, Communications and Applications},
  volume={19},
  number={1},
  pages={1--23},
  year={2023},
  publisher={ACM New York, NY}
}

@inproceedings{wei2022uhd,
  title={Uhd underwater image enhancement via frequency-spatial domain aware network},
  author={Wei, Yiwen and Zheng, Zhuoran and Jia, Xiuyi},
  booktitle={Proceedings of the Asian Conference on Computer Vision},
  pages={299--314},
  year={2022}
}

@article{ren2022reinforced,
  title={Reinforced swin-convs transformer for underwater image enhancement},
  author={Ren, Tingdi and Xu, Haiyong and Jiang, Gangyi and Yu, Mei and Luo, Ting},
  journal={arXiv preprint arXiv:2205.00434},
  year={2022}
}

@inproceedings{ronneberger2015u,
  title={U-net: Convolutional networks for biomedical image segmentation},
  author={Ronneberger, Olaf and Fischer, Philipp and Brox, Thomas},
  booktitle={Medical image computing and computer-assisted intervention--MICCAI 2015: 18th international conference, Munich, Germany, October 5-9, 2015, proceedings, part III 18},
  pages={234--241},
  year={2015},
  organization={Springer}
}

@article{kingma2014adam,
  title={Adam: A method for stochastic optimization},
  author={Kingma, Diederik P and Ba, Jimmy},
  journal={arXiv preprint arXiv:1412.6980},
  year={2014}
}

@inproceedings{wang2003multiscale,
  title={Multiscale structural similarity for image quality assessment},
  author={Wang, Zhou and Simoncelli, Eero P and Bovik, Alan C},
  booktitle={The Thirty-Seventh Asilomar Conference on Signals, Systems \& Computers, 2003},
  volume={2},
  pages={1398--1402},
  year={2003},
  organization={Ieee}
}

@InProceedings{Hu_2022_ACCV,
    author    = {Hu, Zhiqiang and Yu, Tao},
    title     = {Learning to Predict Decomposed Dynamic Filters for Single Image Motion Deblurring},
    booktitle = {Proceedings of the Asian Conference on Computer Vision (ACCV)},
    month     = {December},
    year      = {2022},
    pages     = {4225-4242}
}

@inproceedings{hu2022data,
  title={Data Uncertainty Learning for Single Image Camera Calibration},
  author={Hu, Zhiqiang and Mikuni, Yoshitaka and Arata, Koji},
  booktitle={2022 IEEE 18th International Conference on Automation Science and Engineering (CASE)},
  pages={2140--2147},
  year={2022},
  organization={IEEE}
}


\end{document}